# Information-theoretic Dictionary Learning for Image Classification

Qiang Qiu, *Student Member, IEEE*, Vishal M. Patel, *Member, IEEE*, and Rama Chellappa, *Fellow, IEEE*.

*Abstract*—We present a two-stage approach for learning dictionaries for object classification tasks based on the principle of information maximization. The proposed method seeks a dictionary that is compact, discriminative, and generative. In the first stage, dictionary atoms are selected from an initial dictionary by maximizing the mutual information measure on dictionary compactness, discrimination and reconstruction. In the second stage, the selected dictionary atoms are updated for improved reconstructive and discriminative power using a simple gradient ascent algorithm on mutual information. Experiments using real datasets demonstrate the effectiveness of our approach for image classification tasks.

## I. INTRODUCTION

Sparse signal representations have recently drawn much traction in vision, signal and image processing [1], [2], [3], [4]. This is mainly due to the fact that signals and images of interest can be sparse in some dictionary. Given a redundant dictionary $\mathbf{D}$ and a signal $\mathbf{y}$, finding a sparse representation of $\mathbf{y}$ in $\mathbf{D}$ entails solving the following optimization problem

$$\hat{\mathbf{x}} = \arg\min_{\mathbf{x}} \|\mathbf{x}\|_0 \text{ subject to } \mathbf{y} = \mathbf{D}\mathbf{x}, \quad (1)$$

where the $\ell_0$ sparsity measure $\|\mathbf{x}\|_0$ counts the number of nonzero elements in the vector $\mathbf{x}$. Problem (1) is NP-hard and cannot be solved in a polynomial time. Hence, approximate solutions are usually sought [3], [5], [6], [7].

The dictionary $\mathbf{D}$ can be either based on a mathematical model of the data [3] or it can be trained directly from the data [8]. It has been observed that learning a dictionary directly from training rather than using a predetermined dictionary (such as wavelet or Gabor) usually leads to better representation and hence can provide improved results in many practical applications such as restoration and classification [1], [2], [4], [9].

Various algorithms have been developed for the task of training a dictionary from examples. One of the most commonly used algorithms is the K-SVD algorithm [10]. Given a set of examples $\{\mathbf{y}_i\}_{i=1}^n$, K-SVD finds a dictionary $\mathbf{D}$ that provides the best representation for each example in this set by solving the following optimization problem

$$(\hat{\mathbf{D}}, \hat{\mathbf{X}}) = \arg\min_{\mathbf{D}, \mathbf{X}} \|\mathbf{Y} - \mathbf{D}\mathbf{X}\|_F^2 \text{ subject to } \forall i \ \|\mathbf{x}_i\|_0 \le T_0, \quad (2)$$

Q. Qiu, V. M. Patel, and R. Chellappa are with the Center for Automation Research, UMIACS, University of Maryland, College Park, MD 20742 USA (e-mail: qiu@cs.umd.edu, {pvishalm, rama}@umiacs.umd.edu).

where $\mathbf{x}_i$ represents the $i^{th}$ column of $\mathbf{X}$, $\mathbf{Y}$ is the matrix whose columns are $\mathbf{y}_i$ and $T_0$ is the sparsity parameter. Here, the Frobenius norm is defined as $\|\mathbf{A}\|_F = \sqrt{\sum_{ij} A_{ij}^2}$. The K-SVD algorithm alternates between sparse-coding and dictionary update steps. In the sparse-coding step, $\mathbf{D}$ is fixed and the representation vectors $\mathbf{x}_i$s are found for each example $\mathbf{y}_i$. Then, the dictionary is updated atom-by-atom in an efficient way.

Dictionaries can be trained for both reconstruction and discrimination applications. In the late nineties, Etemand and Chellappa proposed a linear discriminant analysis (LDA) based basis selection and feature extraction algorithm for classification using wavelet packets [11]. Recently, similar algorithms for simultaneous sparse signal representation and discrimination have also been proposed in [12], [13], [14], [15]. Some of the other methods for learning discriminative dictionaries include [16], [17], [18], [19], [20], [21], [12]. Additional techniques may be found within these references.

In this paper, we propose a general method for learning dictionaries for image classification tasks via information maximization. Unlike other previously proposed dictionary learning methods that only consider learning only reconstructive and/or discriminative dictionaries, our algorithm can learn reconstructive, compact and discriminative dictionaries simultaneously. Sparse representation over a dictionary with coherent atoms has the multiple representation problem. A compact dictionary consists of incoherent atoms, and encourages similar signals, which are more likely from the same class, to be consistently described by a similar set of atoms with similar coefficients [21]. A discriminative dictionary encourages signals from different classes to be described by either a different set of atoms, or the same set of atoms but with different coefficients [13], [15], [17]. Both aspects are critical for classification using sparse representation. The additional reconstructive requirement to a compact and discriminative dictionary enhances the robustness of the discriminant sparse representation [13]. All these three criteria are critical for classification using sparse representation.

Our method of training dictionaries consists of two main stages involving greedy atom selection and simple gradient ascent atom updates, resulting in a highly efficient algorithm. In the first stage, dictionary atoms are selected in a greedy way such that the common internal structure of signals belonging to a certain class is extracted while at the same time ensuring global discrimination among the different classes. In the second stage, the dictionary is updated for improved discrimination and reconstruction via a simple gradient ascent method that maximizes the mutual information (MI) between the signals and the dictionary, as well as the sparse coefficients and the class labels.

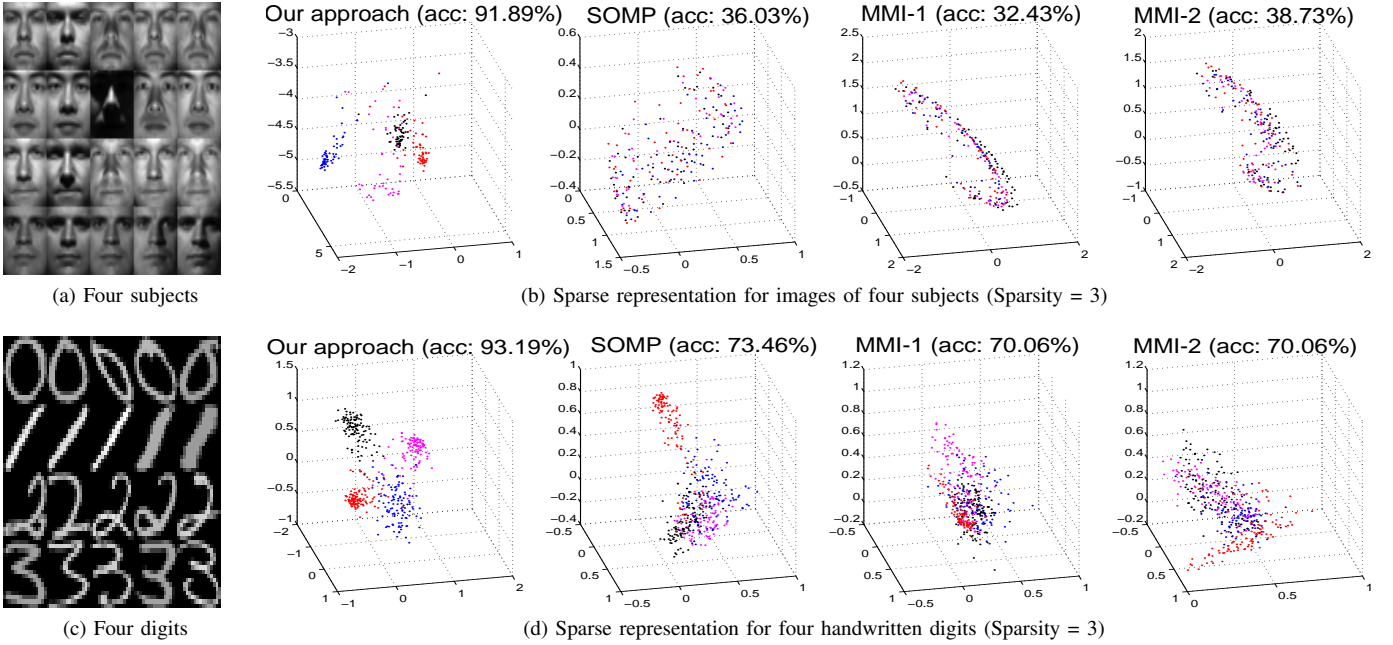

Fig. 1: Sparse representation using dictionaries learned by different approaches (SOMP [22], MMI-1 and MMI-2 [21]). For visualization, sparsity 3 is chosen, i.e., no more than three dictionary atoms are allowed in each sparse decomposition. When signals are represented at once as a linear combination of a common set of atoms, sparse coefficients of all the samples become points in the same coordinate space. Different classes are represented by different colors. The recognition accuracy is obtained through linear SVMs on the sparse coefficients. Our approach provides more discriminative sparse representation which leads to significantly better classification accuracy.

Fig. 1 presents a comparison in terms of the discriminative power of the information-theoretic dictionary learning approach presented in this paper with three state-of-the-art methods. Scatter plots of sparse coefficients obtained using the different methods show that our method provides more discriminative sparse representation, leading to significantly better classification accuracy.

The organization of the paper is as follows. Section II defines and formulates the information theoretic dictionary learning problem. In Section III, the proposed dictionary learning algorithm is detailed. Experimental results are presented in Section IV and Section V concludes the paper with a brief summary and discussion.

## II. BACKGROUND AND PROBLEM FORMULATION

Suppose we are given a set of $N$ signals (images) in an $n$-dim feature space $\mathbf{Y} = [\mathbf{y_1}, ..., \mathbf{y_N}]$, $\mathbf{y_i} \in \mathbb{R}^n$. Given that signals are from $p$ distinct classes and $N_c$ signals are from the $c$-th class, $c \in \{1, \cdots, p\}$, we denote $\mathbf{Y} = \{\mathbf{Y}_c\}_{c=1}^p$, where $\mathbf{Y}_c = [\mathbf{y}_1^c, \cdots, \mathbf{y}_{N_c}^c]$ are signals in the $c$-th class. When the class information is relevant, similarly, we define $\mathbf{X} = \{\mathbf{X}_c\}_{c=1}^p$, where $\mathbf{X}_c = [\mathbf{x}_1^c, \cdots, \mathbf{x}_{N_c}^c]$ is the sparse representation of $\mathbf{Y}_c$.

Given a sample $\mathbf{y}$ at random, the entropy (uncertainty) of the class label in terms of class prior probabilities is defined as

$$H(C) = \sum_c p(c) \left(\frac{1}{p(c)}\right).$$

The mutual information which indicates the decrease in uncertainty about the pattern $\mathbf{y}$ due to the knowledge of the underlying class label $c$ is defined as

$$I(\mathbf{Y}; C) = H(\mathbf{Y}) - H(\mathbf{Y}|C),$$

where $H(\mathbf{Y}|C)$ is the conditional entropy defined as

$$H(\mathbf{Y}|C) = \sum_{\mathbf{y},c} p(\mathbf{y},c) \log\left(\frac{1}{p(\mathbf{y}|c)}\right).$$

Given $\mathbf{Y}$ and an initial dictionary $\mathbf{D}^o$ with $\ell_2$ normalized columns, we aim to learn a compact, reconstructive and discriminative dictionary $\mathbf{D}^*$ via maximizing the mutual information between $\mathbf{D}^*$ and the unselected atoms $\mathbf{D}^o \backslash \mathbf{D}^*$ in $\mathbf{D}^o$, between the sparse codes $\mathbf{X}_{\mathbf{D}^*}$ associated with $\mathbf{D}^*$ and the signal class labels $C$, and finally between the signals $\mathbf{Y}$ and $\mathbf{D}^*$, i.e.,

$$\arg\max_{\mathbf{D}} \lambda_1 I(\mathbf{D}; \mathbf{D}^o\backslash\mathbf{D}) + \lambda_2 I(\mathbf{X_D}; C) + \lambda_3 I(\mathbf{Y}; \mathbf{D}) \quad (3)$$

where $\{\lambda_1, \lambda_2, \lambda_3\}$ are the parameters to balance the contributions from compactness, discriminability and reconstruction terms, respectively.

It is widely known that inclusion of additional criteria, such as a discriminative term, in a dictionary learning framework often involves challenging optimization algorithms [17], [18],

[12]. As discussed above, compactness, discriminability and reconstruction terms are all critical for classification using sparse representation. Maximizing mutual information enables a simple way to unify all three criteria for dictionary learning. As suggested in [23] and [21], maximizing mutual information can also lead to a sub-modular objective function, i.e., a greedy yet near-optimal approach, for dictionary learning.

A two-stage approach is adopted to satisfy (3). In the first stage, each term in (3) is maximized in a unified greedy manner and involves a closed-form evaluation, thus atoms can be greedily selected from the initial dictionary while satisfying (3). In the second stage, the selected dictionary atoms are updated using a simple gradient ascent method to further maximize

$$\lambda_2 I(\mathbf{X_D}; C) + \lambda_3 I(\mathbf{Y}; \mathbf{D}).$$

## III. Information-theoretic Dictionary Learning

In this section, we present the details of our *Information-theoretic Dictionary Learning* (**ITDL**) approach for classification tasks. The dictionary learning procedure is divided into two main steps: *Information-theoretic Dictionary Selection* (**ITDS**) and *Information-theoretic Dictionary Update* (**ITDU**). In what follows, we describe these steps in detail.

### A. Dictionary Selection

Given input signals $\mathbf{Y}$ and an initial dictionary $\mathbf{D}^o$, we select a subset of dictionary atoms $\mathbf{D}^*$ from $\mathbf{D}^o$ via information maximization, i.e., maximizing (3), to encourage the signals from the same class to have very similar sparse representation yet have the discriminative power. In this section, we illustrate why each term in (3) describes the dictionary compactness, discrimination and representation, respectively. We also show that how each term in (3) can be maximized in a unified greedy manner that involves closed-form computations. Therefore, if we start with $\mathbf{D}^* = \emptyset$, and greedily select the next best atom $\mathbf{d}^*$ from $\mathbf{D}^o \backslash \mathbf{D}^*$ which provides an information increase to (3), we obtain a set of dictionary atoms that is compact, reconstructive and discriminative at the same time. To this end, we consider in detail each term in (3) separately.

*1) Dictionary compactness $I(\mathbf{D}^*; \mathbf{D}^o \backslash \mathbf{D}^*)$:* The dictionary compactness $I(\mathbf{D}^*; \mathbf{D}^o \backslash \mathbf{D}^*)$ has been studied in our early work [21]. We summarize [21] to complete our information-driven dictionary selection discussion. [21] suggests dictionary compactness is required to avoid the multiple sparse representation problem for better classification performance. In [21], we first model sparse representation through a Gaussian Process model to define the mutual information $I(\mathbf{D}^*; \mathbf{D}^o \backslash \mathbf{D}^*)$. A compact dictionary can be then obtained as follows: we start with $\mathbf{D}^* = \emptyset$ and iteratively choose the next best dictionary item $\mathbf{d}^*$ from $\mathbf{D}^o \backslash \mathbf{D}^*$ which provides a maximum increase in mutual information, i.e.,

$$\arg\max\nolimits_{\mathbf{d}^* \in \mathbf{D}^o \backslash \mathbf{D}^*} I(\mathbf{D}^* \cup \mathbf{d}^*; \mathbf{D}^o \backslash (\mathbf{D}^* \cup \mathbf{d}^*)) - I(\mathbf{D}^*; \mathbf{D}^o \backslash \mathbf{D}^*). \tag{4}$$

It has been proved in [23] that the above greedy algorithm serves a polynomial-time approximation that is within $(1 - 1/e)$ of the optimum.

*2) Dictionary Discrimination $I(\mathbf{X_{D^*}}; C)$:* Using any pursuit algorithm such as OMP [6], we initialize the sparse coefficients $\mathbf{X_{D^o}}$ for input signals $\mathbf{Y}$ and an initial dictionary $\mathbf{D}^o$. Given $\mathbf{X_{D^*}}$ are sparse coefficients associated with the desired set of atoms $\mathbf{D}^*$ and $C$ are the class labels for input signals $\mathbf{Y}$, based on [24], an upper bound on the Bayes error over sparse representation $E(\mathbf{X_{D^*}})$ is obtained as

$$\frac{1}{2}(H(C) - I(\mathbf{X_{D^*}}; C)).$$

This bound is minimized when $I(\mathbf{X_{D^*}}; C)$ is maximized. Thus, a discriminative dictionary $\mathbf{D}^*$ is obtained via

$$\arg\max_{\mathbf{D}^*} I(\mathbf{X_{D^*}}; C). \tag{5}$$

We maximize (5) using a greedy algorithm initialized by $\mathbf{D}^* = \emptyset$ and iteratively choosing the next best dictionary atom $\mathbf{d}^*$ from $\mathbf{D}^o \backslash \mathbf{D}^*$ which provides a maximum mutual information increase, i.e.,

$$\arg\max_{\mathbf{d}^* \in \mathbf{D}^o \backslash \mathbf{D}^*} I(\mathbf{X_{D^* \cup d^*}}; C) - I(\mathbf{X_{D^*}}; C), \tag{6}$$

where $I(\mathbf{X_{D^*}}; C)$ is evaluated as follows

$$I(\mathbf{X_{D^*}}; C) = H(\mathbf{X_{D^*}}) - H(\mathbf{X_{D^*}}|C) \tag{7}$$
$$= H(\mathbf{X_{D^*}}) - \sum_{c=1}^{p} p(c) H(\mathbf{X_{D^*}}|c).$$

Entropy measures in (7) involve computation of probability density functions $p(\mathbf{X_{D^*}})$ and $p(\mathbf{X_{D^*}}|c)$. We adopt the kernel density estimation method [25] to non-parametrically estimate the probability densities. Using isotropic Gaussian kernels (i.e. $\mathbf{\Sigma} = \sigma^2 \mathbf{I}$, where $\mathbf{I}$ is the identity matrix), the class dependent density for the $c$-th class can be estimated as

$$p(\mathbf{x}|c) = \frac{1}{N_c} \sum_{j=1}^{N_c} \mathbb{K}_G(\mathbf{x} - \mathbf{x}_j^c, \sigma^2 \mathbf{I}), \tag{8}$$

where $\mathbb{K}_G$ is a $d$-dim Gaussian kernel defined as

$$\mathbb{K}_G(\mathbf{x}, \mathbf{\Sigma}) = \frac{1}{(2\pi)^{\frac{d}{2}} |\mathbf{\Sigma}|^{\frac{1}{2}}} \exp\left(-\frac{1}{2} \mathbf{x}^T \mathbf{\Sigma}^{-1} \mathbf{x}\right). \tag{9}$$

With $p(c) = \frac{N_c}{N}$, we can estimate $p(\mathbf{x})$ as

$$p(\mathbf{x}) = \sum_c p(\mathbf{x}|c) p(c).$$

*3) Dictionary Representation $I(\mathbf{Y}; \mathbf{D}^*)$:* A representative dictionary $\mathbf{D}^*$ maximizes the mutual information between dictionary atoms and the signals, i.e.,

$$\arg\max_{\mathbf{D}^*} I(\mathbf{Y}; \mathbf{D}^*). \tag{10}$$

We obtain a representative dictionary via a similar greedy manner as discussed above. That is, we iteratively choose the next best dictionary atom $\mathbf{d}^*$ from $\mathbf{D}^o \backslash \mathbf{D}^*$ which provides the maximum increase in mutual information,

$$\arg\max_{\mathbf{d}^* \in \mathbf{D}^o \backslash \mathbf{D}^*} I(\mathbf{Y}; \mathbf{D}^* \cup \mathbf{d}^*) - I(\mathbf{Y}; \mathbf{D}^*). \tag{11}$$

By assuming the signals are drawn independently and using the chain-rule of entropies, we can evaluate $I(\mathbf{Y}; \mathbf{D}^*)$ as

$$I(\mathbf{Y}; \mathbf{D}^*) = H(\mathbf{Y}) - H(\mathbf{Y}|\mathbf{D}^*) \quad (12)$$
$$= H(\mathbf{Y}) - \sum_{i=1}^{N} H(\mathbf{y}_i|\mathbf{D}^*).$$

$H(\mathbf{Y})$ is independent of dictionary selection and can be ignored. To evaluate $H(\mathbf{y}_i|\mathbf{D}^*)$ in (12), we define $p(\mathbf{y}_i|\mathbf{D}^*)$ through the following relation holding for each input signal $\mathbf{y}_i$,

$$\mathbf{y}_i = \mathbf{D}^*\mathbf{x}_i + \mathbf{r}_i,$$

where $\mathbf{r}_i$ is a Gaussian residual vector with variance $\sigma_r^2$. Such a relation can be written in a probabilistic form as,

$$p(\mathbf{y}_i|\mathbf{D}^*) \propto \exp(-\frac{1}{2\sigma_r^2}||\mathbf{y}_i - \mathbf{D}^*\mathbf{x}_i||^2).$$

*4) Selection of $\lambda_1$, $\lambda_2$ and $\lambda_3$:* The parameters $\lambda_1$, $\lambda_2$ and $\lambda_3$ in (3) are data dependent and can be estimated as the ratio between the maximal information gained from an atom to the respective compactness, discrimination and reconstruction measure, i.e.,

$$\lambda_1 = 1, \quad (13)$$
$$\lambda_2 = \frac{\max_i I(\mathbf{X}_{\mathbf{d}_i}; C)}{\max_i I(\mathbf{d}_i; \mathbf{D}^o \backslash \mathbf{d}_i)},$$
$$\lambda_3 = \frac{\max_i I(\mathbf{Y}; \mathbf{d}_i)}{\max_i I(\mathbf{d}_i; \mathbf{D}^o \backslash \mathbf{d}_i)}.$$

For each term in (3), only the first greedily selected atom based on (4), (6) and (11), respectively are involved in parameter estimation. This leads to an efficient process in finding parameters.

### B. Dictionary Update

A representative and discriminative dictionary $\mathbf{D}$ produces the maximal MI between the sparse coefficients and the class labels, as well as the signals and the dictionary, i.e.,

$$\max_{\mathbf{D}} \lambda_2 I(\mathbf{X}_{\mathbf{D}}; C) + \lambda_3 I(\mathbf{Y}; \mathbf{D}).$$

In the dictionary update stage, we update the set of selected dictionary atoms $\mathbf{D}$ to further enhance the discriminability and representation.

To achieve sparsity, we assume the cardinality of the set of selected atoms $\mathbf{D}$ is much smaller than the dimension of the signal feature space. Under such an assumption, the sparse representation of signals $\mathbf{Y}$ can be obtained as $\mathbf{X}_{\mathbf{D}} = \mathbf{D}^\dagger \mathbf{Y}$ which minimizes the representation error $||\mathbf{Y} - \mathbf{D}\mathbf{X}_{\mathbf{D}}||_F^2$, where

$$\mathbf{D}^\dagger = (\mathbf{D}^\mathbf{T}\mathbf{D})^{-1}\mathbf{D}^\mathbf{T}.$$

Thus, updating dictionary atoms for improving discriminability while maintaining representation is transformed into finding $\mathbf{D}^\dagger$ that maximizes

$$I(\mathbf{D}^\dagger \mathbf{Y}; C).$$

*1) A Differentiable Objective Function:* To enable a simple gradient ascent method for dictionary update, we first approximate $I(\mathbf{D}^\dagger \mathbf{Y}; C)$ using a differentiable objective function. $I(\mathbf{X}; C)$ can be viewed as the Kullback-Leibler (KL) divergence $D(p||q)$ between $p(\mathbf{X}, C)$ and $p(\mathbf{X})p(C)$, where $\mathbf{X} = \mathbf{D}^\dagger \mathbf{Y}$. Motivated by [26], we approximate the KL divergence $D(p||q)$ with the quadratic divergence (QD), defined as

$$Q(p||q) = \int_t (p(t) - q(t))^2 \, dt,$$

making $I(\mathbf{X}; C)$ differentiable. Due to the property that

$$D(p||q) \geq \frac{1}{2}Q(p||q),$$

by maximizing the QD, one can also maximize a lower bound to the KL divergence. With QD, $I(\mathbf{X}; C)$ can now be evaluated as,

$$I_Q(\mathbf{X}; C) = \sum_c \int_{\mathbf{x}} p(\mathbf{x}, c)^2 \, d\mathbf{x} \quad (14)$$
$$- 2\sum_c \int_{\mathbf{x}} p(\mathbf{x}, c)p(\mathbf{x})p(c) \, d\mathbf{x}$$
$$+ \sum_c \int_{\mathbf{x}} p(\mathbf{x})^2 p(c)^2 \, d\mathbf{x}.$$

In order to evaluate the individual terms in (14), we need to derive expressions for the kernel density estimates of various density terms appearing in (14). Observe that for the two Gaussian kernels in (9), the following holds

$$\int_{\mathbf{x}} \mathbb{K}_G(\mathbf{x}-\mathbf{s}_i, \mathbf{\Sigma}_1)\mathbb{K}_G(\mathbf{x}-\mathbf{s}_j, \mathbf{\Sigma}_2) \, d\mathbf{x} = \mathbb{K}_G(\mathbf{s}_i-\mathbf{s}_j, \mathbf{\Sigma}_1+\mathbf{\Sigma}_2). \quad (15)$$

Using (8), $p(c) = \frac{N_c}{N}$ and $p(\mathbf{x}, c) = p(\mathbf{x}|c)p(c)$, we have

$$p(\mathbf{x}, c) = \frac{1}{N} \sum_{j=1}^{N_c} \mathbb{K}_G(\mathbf{x} - \mathbf{x}_j^c, \sigma^2 \mathbf{I}).$$

Similarly, since $p(\mathbf{x}) = \sum_c p(\mathbf{x}, c)$, we have

$$p(\mathbf{x}) = \frac{1}{N} \sum_{i=1}^{N} \mathbb{K}_G(\mathbf{x} - \mathbf{x}_i, \sigma^2 \mathbf{I}).$$

Inserting expressions for $p(\mathbf{x}, c)$ and $p(\mathbf{x})$ into (14) and using (15), we get the following closed form

$$I_Q(\mathbf{X}; C) = \frac{1}{N^2} \sum_{c=1}^{p} \sum_{k=1}^{N_c} \sum_{l=1}^{N_c} \mathbb{K}_G(\mathbf{x}_k^c - \mathbf{x}_l^c, 2\sigma^2 \mathbf{I}) \quad (16)$$
$$- \frac{2}{N^2} \sum_{c=1}^{p} \frac{N_c}{N} \sum_{j=1}^{N_c} \sum_{k=1}^{N} \mathbb{K}_G(\mathbf{x}_j^c - \mathbf{x}_k, 2\sigma^2 \mathbf{I})$$
$$+ \frac{1}{N^2} \left( \sum_{c=1}^{p} \left(\frac{N_c}{N}\right)^2 \right) \sum_{k=1}^{N} \sum_{l=1}^{N} \mathbb{K}_G(\mathbf{x}_k - \mathbf{x}_l, 2\sigma^2 \mathbf{I}). \quad (17)$$

*2) Gradient Ascent Update:* For simplicity, we define a new matrix $\Phi$ as

$$\Phi \triangleq (\mathbf{D}^\dagger)^{\mathbf{T}}.$$

Once we have estimated $I_Q(\mathbf{X}; C)$ as a function of the data set in a differential form, where $\mathbf{X} = \Phi^T \mathbf{Y}$, we can use gradient ascent on $I_Q(\mathbf{X}; C)$ to search for the optimal $\Phi$ maximizing the quadratic mutual information with

$$\Phi_{k+1} = \Phi_k + \nu \frac{\partial I_Q}{\partial \Phi}|_{\Phi = \Phi_k}$$

where $\nu \geq 0$ defining the step size, and

$$\frac{\partial I_Q}{\partial \Phi} = \sum_{c=1}^{p} \sum_{i=1}^{N_c} \frac{\partial I_Q}{\partial \mathbf{x}_i^c} \frac{\partial \mathbf{x}_i^c}{\partial \Phi}.$$

Since $\mathbf{x}_i^c = \Phi^T \mathbf{y}_i^c$, we get

$$\frac{\partial \mathbf{x}_i^c}{\partial \Phi} = (\mathbf{y}_i^c)^T.$$

Note that

$$\frac{\partial}{\partial \mathbf{x}_i} \mathbb{K}_G(\mathbf{x}_i - \mathbf{x}_j, 2\sigma^2 \mathbf{I}) = \mathbb{K}_G(\mathbf{x}_i - \mathbf{x}_j, 2\sigma^2 \mathbf{I}) \frac{(\mathbf{x}_i - \mathbf{x}_j)}{2\sigma^2}.$$

We have

$$\frac{\partial}{\partial \mathbf{x}_i^c} I_Q = \frac{1}{N^2 \sigma^2} \sum_{k=1}^{N_c} \mathbb{K}_G(\mathbf{x}_k^c - \mathbf{x}_i^c, 2\sigma^2 \mathbf{I})(\mathbf{x}_k^c - \mathbf{x}_i^c)$$
$$- \frac{2}{N^2 \sigma^2} \left(\sum_{c=1}^{p} \left(\frac{N_c}{N}\right)^2\right) \sum_{k=1}^{N} \mathbb{K}_G(\mathbf{x}_k - \mathbf{x}_i^c, 2\sigma^2 \mathbf{I})(\mathbf{x}_k - \mathbf{x}_i^c)$$
$$+ \frac{1}{N^2 \sigma^2} \sum_{k=1}^{p} \frac{N_k + N_c}{2N} \sum_{j=1}^{N_k} \mathbb{K}_G(\mathbf{x}_j^k - \mathbf{x}_i^c, 2\sigma^2 \mathbf{I})(\mathbf{x}_j^k - \mathbf{x}_i^c).$$
(18)

Once $\Phi$ is updated, the dictionary $\mathbf{D}$ can be updated using the relation $\Phi = (\mathbf{D}^\dagger)^{\mathbf{T}}$. Such dictionary updates guarantee convergence to a local maximum due to the fact that the quadratic divergence is bounded [27].

### C. Dictionary Learning Framework

Given a dictionary $\mathbf{D}^o$, a set of signals $\mathbf{Y}$, the class labels $C$ and a sparsity level $T$, the supervised sparse coding method given in Algorithm 1 represents these signals at once as a linear combination of a common subset of $T$ atoms in $\mathbf{D}$, where $T$ is much smaller than the dimension of the signal feature space to achieve sparsity. We obtain a sparse representation as each signal has no more than $T$ coefficients in its decomposition. The advantage of simultaneous sparse decomposition for classification has been discussed in [13]. Such simultaneous decompositions extract the internal structure of given signals and neglects minor intra-class variations. The ITDS stage in Algorithm 1 ensures such common set of atoms are compact, discriminative and reconstructive.

When the internal structures of signals from different classes can not be well represented in a common linear subspace, Algorithm 2 illustrates supervised sparse coding with a dedicated set of atoms per class. It is noted in Algorithm 2 that both the discriminative and reconstructive terms in ITDS are handled on a class by class basis.

**Input**: Dictionary $\mathbf{D}^o$, signals $\mathbf{Y}$, class labels $C$, sparsity level $T$
**Output**: sparse coefficients $\mathbf{X}$, reconstruction $\hat{\mathbf{Y}}$
**begin**
  **Initialization stage:**
  1. Initialize $\mathbf{X}$ with any pursuit algorithm,
  $i = 1, \cdots, N \quad \min_{\mathbf{x}_i} \|\mathbf{y}_i - \mathbf{D}^o \mathbf{x_i}\|_2^2 \ s.t. \ \|\mathbf{x}_i\|_0 \leq T$.
  **ITDS stage (shared atoms):**
  2. Estimate $\lambda_1$, $\lambda_2$ and $\lambda_3$ from $\mathbf{Y}$, $\mathbf{X}$ and $C$;
  3. Find $T$ most compact, discriminative and reconstructive atoms:
  $\mathbf{D}^* \leftarrow \emptyset; \Gamma \leftarrow \emptyset$ ;
  **for** *t=1 to T* **do**
    $\mathbf{d}^* \leftarrow \arg\max_{\mathbf{d} \in \mathbf{D}^o \setminus \mathbf{D}^*} \lambda_1[I(\mathbf{D}^* \cup \mathbf{d}; \mathbf{D}^o \setminus (\mathbf{D}^* \cup \mathbf{d})) - I(\mathbf{D}^*; \mathbf{D}^o \setminus \mathbf{D}^*)] + \lambda_2[I(\mathbf{X}_{\mathbf{D}^* \cup \mathbf{d}}; C) - I(\mathbf{X}_{\mathbf{D}^*}; C)] + \lambda_3[I(\mathbf{Y}; \mathbf{D}^* \cup \mathbf{d}) - I(\mathbf{Y}; \mathbf{D}^*)]$;
    $\mathbf{D}^* \leftarrow \mathbf{D}^* \bigcup \mathbf{d}^*$;
    $\Gamma \leftarrow \Gamma \bigcup \gamma^*$, $\gamma^*$ is the index of $\mathbf{d}^*$ in $\mathbf{D}^o$ ;
  **end**
  4. Compute sparse codes and reconstructions:
    $\mathbf{X} \leftarrow pinv(\mathbf{D}^*)\mathbf{Y}$;
    $\hat{\mathbf{Y}} \leftarrow \mathbf{D}^* \mathbf{X}$;
  5. return $\mathbf{X}$, $\hat{\mathbf{Y}}$, $\mathbf{D}^*$, $\Gamma$ ;
**end**

**Algorithm 1:** Sparse coding with global atoms.

**Input**: Dictionary $\mathbf{D}^o$, signals $\mathbf{Y} = \{\mathbf{Y}_c\}_{c=1}^p$, sparsity level $T$
**Output**: sparse coefficients $\{\mathbf{X}_c\}_{c=1}^p$, reconstruction $\{\hat{\mathbf{Y}}_c\}_{c=1}^p$
**begin**
  **Initialization stage:**
  1. Initialize $\mathbf{X}$ with any pursuit algorithm,
  $i = 1, \cdots, N \quad \min_{\mathbf{x}_i} \|\mathbf{y}_i - \mathbf{D}^o \mathbf{x_i}\|_2^2 \ s.t. \ \|\mathbf{x}_i\|_0 \leq T$.
  **ITDS stage (dedicated atoms):**
  **for** *c=1 to p* **do**
    2. $C_c \leftarrow \{c_i | c_i = 1 \text{ if } y_i \in \mathbf{Y}_c, 0 \text{ otherwise }\}$ ;
    3. Estimate $\lambda_1$, $\lambda_2$ and $\lambda_3$ from $\mathbf{Y}_c$, $\mathbf{X}$ and $C_c$;
    4. Find $T$ most compact, discriminative and reconstructive atoms for class $c$:
    $\mathbf{D}^* \leftarrow \emptyset; \Gamma \leftarrow \emptyset$ ;
    **for** *t=1 to T* **do**
      $\mathbf{d}^* \leftarrow \arg\max_{\mathbf{d} \in \mathbf{D}^o \setminus \mathbf{D}^*} \lambda_1[I(\mathbf{D}^* \cup \mathbf{d}; \mathbf{D}^o \setminus (\mathbf{D}^* \cup \mathbf{d})) - I(\mathbf{D}^*; \mathbf{D}^o \setminus \mathbf{D}^*)] + \lambda_2[I(\mathbf{X}_{\mathbf{D}^* \cup \mathbf{d}}; C_c) - I(\mathbf{X}_{\mathbf{D}^*}; C_c)] + \lambda_3[I(\mathbf{Y}_c; \mathbf{D}^* \cup \mathbf{d}) - I(\mathbf{Y}_c; \mathbf{D}^*)]$;
      $\mathbf{D}^* \leftarrow \mathbf{D}^* \bigcup \mathbf{d}^*$;
      $\Gamma \leftarrow \Gamma \bigcup \gamma^*$, $\gamma^*$ is the index of $\mathbf{d}^*$ in $\mathbf{D}^o$ ;
    **end**
    $\mathbf{D}_c^* \leftarrow \mathbf{D}^*; \Gamma_c \leftarrow \Gamma$;
    5. Compute sparse codes and reconstructions:
      $\mathbf{X}_c \leftarrow pinv(\mathbf{D}_c^*)\mathbf{Y}_c$;
      $\hat{\mathbf{Y}}_c \leftarrow \mathbf{D}_c^* \mathbf{X}_c$;
  **end**
  6. return $\{\mathbf{X}_c\}_{c=1}^p$, $\{\hat{\mathbf{Y}}_c\}_{c=1}^p$, $\{\mathbf{D}_c^*\}_{c=1}^p$, $\{\Gamma_c\}_{c=1}^p$ ;
**end**

**Algorithm 2:** Sparse coding with atoms per class.

A sparse dictionary learning framework, such as K-SVD [10] which learns a dictionary that minimizes the reconstruction error, usually consists of sparse coding and update stages. In K-SVD, at the coding stage, a pursuit algorithm is employed to select a set of atoms for each signal; and at the update stage,

```
Input: Dictionary D°, signals Y = {Y_c}_{c=1}^p, class labels C,
       sparsity level T, update step ν
Output: Learned dictionary D, sparse coefficients X,
        reconstruction Ŷ
begin
    Sparse coding stage:
    Use supervised sparse coding to obtain {D*_c}_{c=1}^p.
    ITDU stage:
    foreach class c do
        [In the shared atom case, use the global label C
        instead of C_c, and one iteration is required as the same
        D*_c is used for all classes.]
        C_c ← {c_i | c_i = 1 if y_i ∈ Y_c, 0 otherwise };
        Φ_1 ← pinv(D*_c)^T;
        X ← pinv(D*_c)Y;
        repeat
            Φ_{k+1} = Φ_k + ν ∂I_Q(X,C_c)/∂Φ |_{Φ=Φ_k} ;
            D* ← pinv(Φ_{k+1}^T);
            X ← pinv(D*)Y;
        until convergence;
        D*_c ← D* ;
    end
    foreach class c do
        X_c ← pinv(D*_c)Y_c;
        Ŷ_c ← D*_c X_c;
    end
    return {X_c}_{c=1}^p, {Ŷ_c}_{c=1}^p, {D*_c}_{c=1}^p ;
end
```

**Algorithm 3:** Sparse coding with atom updates.

the selected atoms are updated through SVD for improved reconstruction. Similarly, in Algorithm 3, at the coding stage, ITDS is employed to select a set of atoms for each class of signals; and at the update stage, the selected atoms are updated through ITDU for improved reconstruction and discrimination. Algorithm 3 is also applicable to the case when sparse coding is achieved using global atoms.

## IV. EXPERIMENTAL EVALUATION

This section presents an experimental evaluation on three public datasets: the Extended YaleB face dataset [28], the USPS handwritten digits dataset [29], and the 15-Scenes dataset [30]. The Extended YaleB dataset contains 2414 frontal face images for 38 individuals. This dataset is challenging due to varying illumination conditions and expressions. The USPS dataset consists of 8-bit 16×16 images of "0" through "9" and 1100 examples for each class. The 15-Scenes dataset contains 4485 images falling into 15 scene categories. The 15 categories include images of living rooms, kitchens, streets, industrials, etc.. In all of our experiments, linear SVMs on the sparse coefficients are used for classifiers. First, we thoroughly evaluate the basic behaviors of the proposed dictionary learning method. Then we evaluate the discriminative power of the ITDL dictionary over the full Extended YaleB dataset, the full USPS dataset, and the 15-Scenes dataset.

### A. Evaluation with Illustrative Examples

To enable visualized illustrations, we conduct the first set of experiments on the first four subjects in the Extended YaleB face dataset and the first four digits in the USPS digit dataset. Half of the data are used for training and the rest is used for testing.

*1) Comparing Atom Selection Methods:* We initialize a 128 sized dictionary using the K-SVD algorithm [10] on the training face images of the first four subjects in the Extended YaleB dataset. A K-SVD dictionary only minimizes the reconstruction error and is not yet optimal for classification tasks. Though one can also initialize the dictionary directly with training samples or even with random noise, a better initial dictionary generally helps ITDL in terms of classification performance, due to the fact that an ITDL dictionary converges to a local maximum.

In Fig. 2, we present the recognition accuracy and the reconstruction error with different sparsity on the first four subjects in the Extended YaleB dataset. The Root Mean Square Error (RMSE) is employed to measure the reconstruction error. To illustrate the impact of the compactness, discrimination and reconstruction terms in (3), we keep one term at a time for the three selection approaches, i.e., the compact, the discriminative and the reconstructive method. The compact method is equivalent to MMI-1 [21].

Parameters $\lambda_1$, $\lambda_2$ and $\lambda_3$ in (3) are estimated as discussed in Section III-A4. As the dictionary learning criteria becomes less critical when sparsity increases, i.e., more energies in signals are actually preserved, we focus on curves in Fig. 2 when sparsity<20. Although sparse coding methods generally perform well for face recognition, it is still easy to notice that the proposed ITDS method using all three terms (red) significantly outperforms those which optimize just one of the three terms, compactness (black), discrimination (blue), and representation (green), in terms of recognition accuracy. For example, the discrimination term alone (blue) leads to a better initial but poor overall recognition performance. The proposed ITDS method also provides moderate reconstruction error.

It is noted that IDS exhibits comparable recognition accuracy to MMI-2 (pink) [21] with global atoms, and significantly outperforms it with class dedicated atoms. The reason is that, instead of explicitly considering the discriminability of dictionary atoms, MMI-2 enforces the diversity of classes associated with atoms. Such class diversity criteria becomes less effective when there are only two classes in the dedicate atom case. In Fig. 2, it is interesting to note that the reconstructive method delivers nearly identical recognition accuracy and RMSE to SOMP [22] with both the shared and dedicated atoms, given the different formulations of two methods. The proposed dictionary selection using all three terms provides a good local optimum to converge at the dictionary update stage.

*2) Enhanced Discriminability with Atom Update:* We illustrate how the discriminability of dictionary atoms selected by the ITDS method can be further enhanced using the proposed ITDU method. We initialize a 128 sized K-SVD dictionary for the face images and a 64 sized K-SVD dictionary for the the digit images. Sparsity 2 is adopted for visualization, as the non-zero sparse coefficients of each image can now be plotted as a 2-D point. In Fig. 3, with a common set of atoms shared over all classes, sparse coefficients of all samples become points in the same 2-D coordinate space. Different classes are represented by different colors. The original images are also shown and placed at the coordinates defined by their non-zero sparse coefficients. The atoms to be updated in Fig. 3a and 3d are selected using ITDS. We can see from Fig. 3 that the proposed ITDU method makes sparse coefficients of different classes more discriminative, leading to significantly improved classification accuracy. Fig. 4 shows that the ITDU method also enhances the discriminability of atoms dedicated to each class. It is noted that, though the dictionary update sometimes only converges after a considerable number of iterations, based on our experience, the first 50 to 100 iterations in general bring significant improvement in classification accuracy.

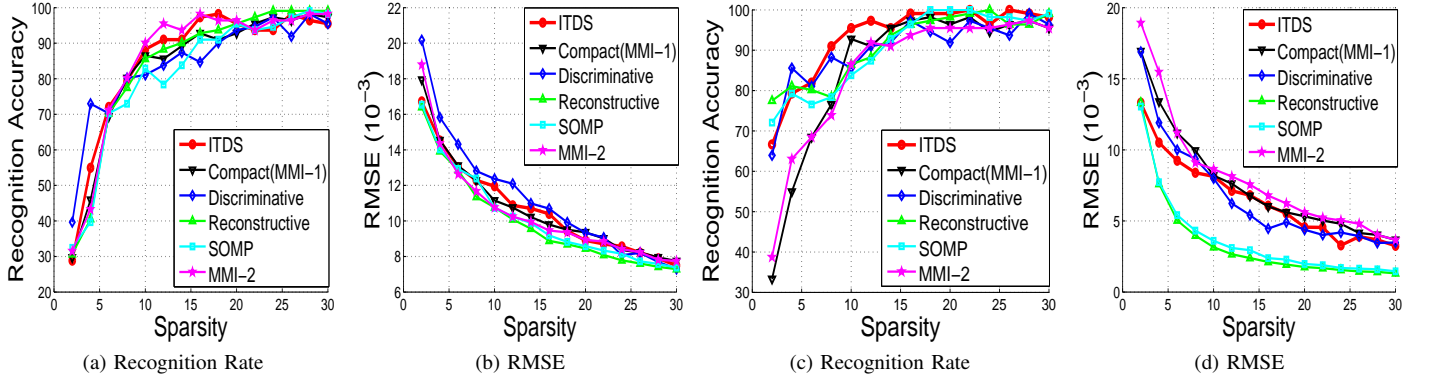

Fig. 2: Recognition accuracy and RMSE on the YaleB dataset using different dictionary selection methods. We vary the sparsity level, i.e., the maximal number of dictionary atoms that are allowed in each sparse decomposition. In (a) and (b), a global set of common atoms are selected for all classes. In (c) and (d), a dedicated set of atoms are selected per class. In both cases, the proposed ITDS (red lines) provides the best recognition performance and moderate reconstruction error.

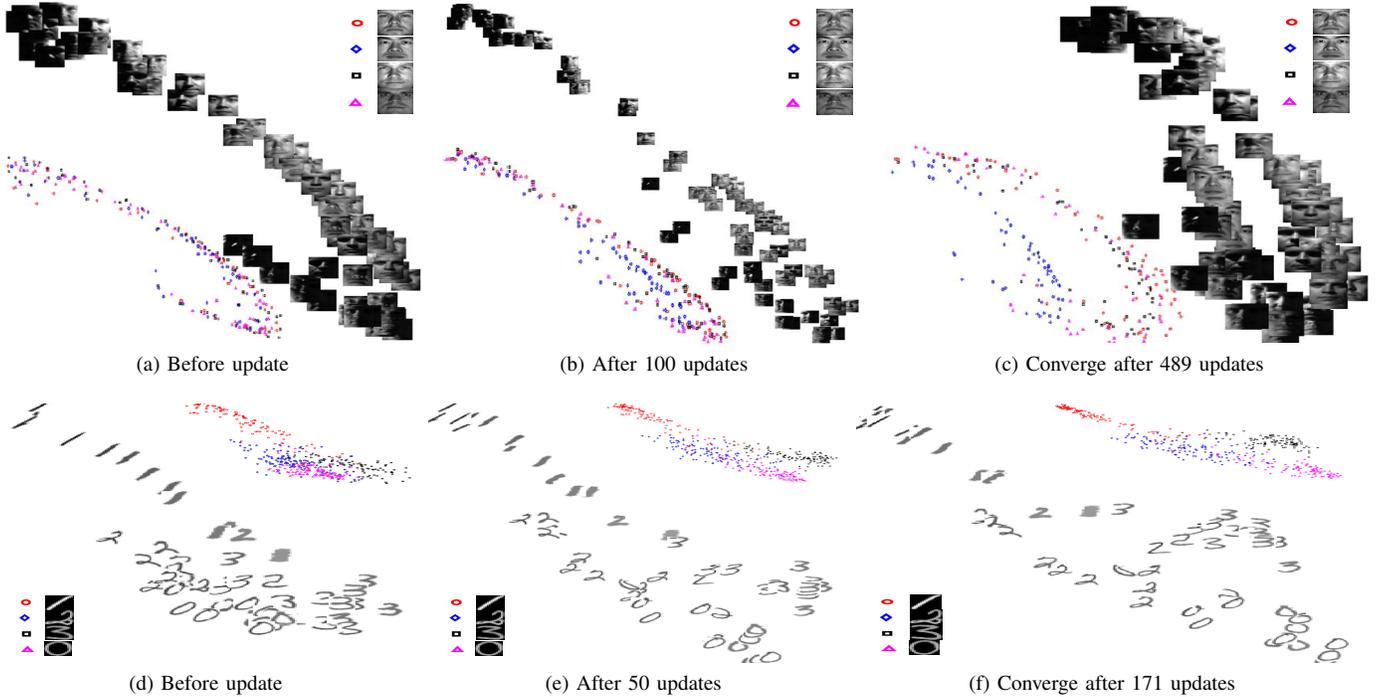

Fig. 3: Information-theoretic dictionary update with global atoms shared over classes. For a better visual representation, sparsity 2 is chosen and a randomly selected subset of all samples are shown. The recognition rate associated with (a), (b), and (c) are: $30.63\%$, $42.34\%$ and $51.35\%$. The recognition rate associated with (d), (e), and (f) are: $73.54\%$, $84.45\%$ and $87.75\%$. Note that the proposed ITDU effectively enhances the discriminability of the set of common atoms.

*3) Enhanced Reconstruction with Atom Update:* From Fig. 5e, we notice obvious errors in the reconstructed digits, shown in Fig. 5d with atoms selected from the initial K-SVD dictionary using ITDS. After 30 ITDU iterations, Fig. 5f shows that all digits are reconstructed correctly with a unified intra-class structure and limited intra-class variation. This leads to a more accurate classification as shown in Fig. 4. It is noted that Fig. 5 and Fig. 4 are results from the same set of experiments.

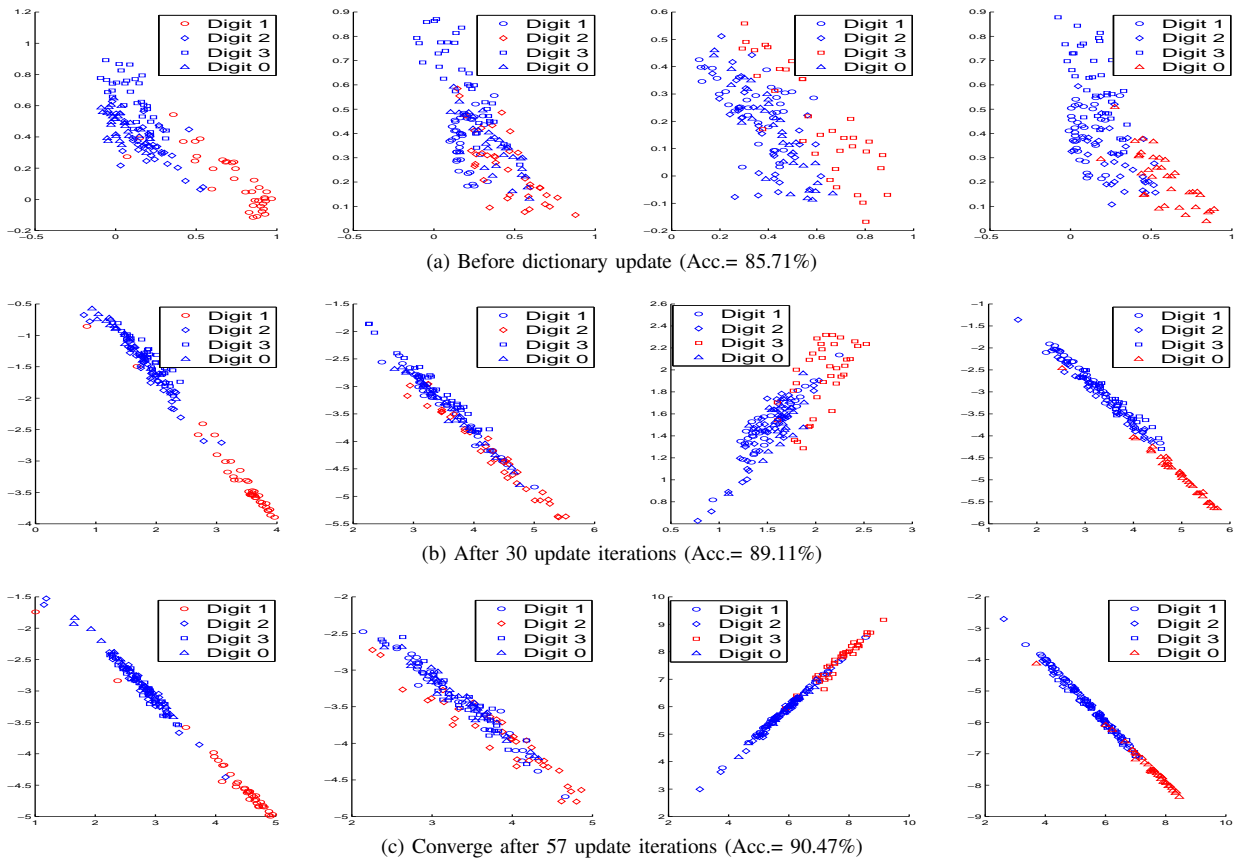

Fig. 4: Information-theoretic dictionary update with dedicated atoms per class. The first four digits in the USPS digit dataset are used. Sparsity 2 is chosen for visualization. In each figure, signals are first represented at once as a linear combination of the dedicated atoms for the class colored by red, then sparse coefficients of all signals are plotted in the same 2-D coordinate space. The proposed ITDU effectively enhances the discriminability of the set of dedicated atoms.

TABLE I: Classification rate (%) on the USPS dataset.

| Proposed | SDL-D [18] | SRSC [15] | FDDL [12] | k-NN | SVM-Gauss |
|---|---|---|---|---|---|
| **98.28** | 96.44 | 93.95 | 96.31 | 94.80 | 95.80 |

TABLE II: Classification rate (%) on the 15 scenes dataset.

| Proposed | ScSPM [31] | KSPM [30] | KC [32] | LSPM [31] |
|---|---|---|---|---|
| **81.13** | 80.28 | 76.73 | 76.67 | 65.32 |

TABLE III: Classification rate (%) on the Extended YaleB face dataset.

| Proposed | D-KSVD [19] | LC-KSVD [20] | K-SVD [10] | SRC [33] | LLC [34] |
|---|---|---|---|---|---|
| **95.39** | 94.10 | 95.00 | 93.1 | 80.5 | 90.7 |

As can be seen from Fig. 5g, after ITDU converges, all digits are reconstructed correctly with the true underlying intra-class structures, i.e., the left-slanted and right-slanted styles for both digits "1" and "0". Fig. 5h shows the images in Fig. 5d with 60% missing pixels. The recognition rate for Fig. 5i, Fig. 5j, and Fig. 5k are 76.87%, 85.03% and 85.71%, respectively.

### B. Discriminability of ITDL Dictionaries

We evaluate the discriminative power of ITDL dictionaries over the complete USPS dataset, where we use 7291 images for training and 2007 images for testing, and the Extended YaleB face dataset, where we randomly select half of the images as training and the other half for testing, and finally the 15-Scenes dataset, where we randomly use 100 images per class for training and used the remaining data for testing.

For each dataset, we initialize a 512 sized dictionary from K-SVD and set the sparsity to be 30. Then we perform 30 iterations of dictionary update and report the peak classification performance. Here we adopt a dedicated set of atoms for each class and input the concatenated sparse representation into a linear SVM classifier. For the Extended YaleB face dataset, we adopt the same experimental setup in [20]. As shown in Table I, Table II, and Table III, our method is comparable to some of the competitive discriminative dictionary learning algorithms such as SDL-D [18], SRSC [15], D-KSVD [19] and LC-KSVD [20]. Note that, our method is flexible enough that it can be applied over any dictionary learning schemes to enhance the discriminability.

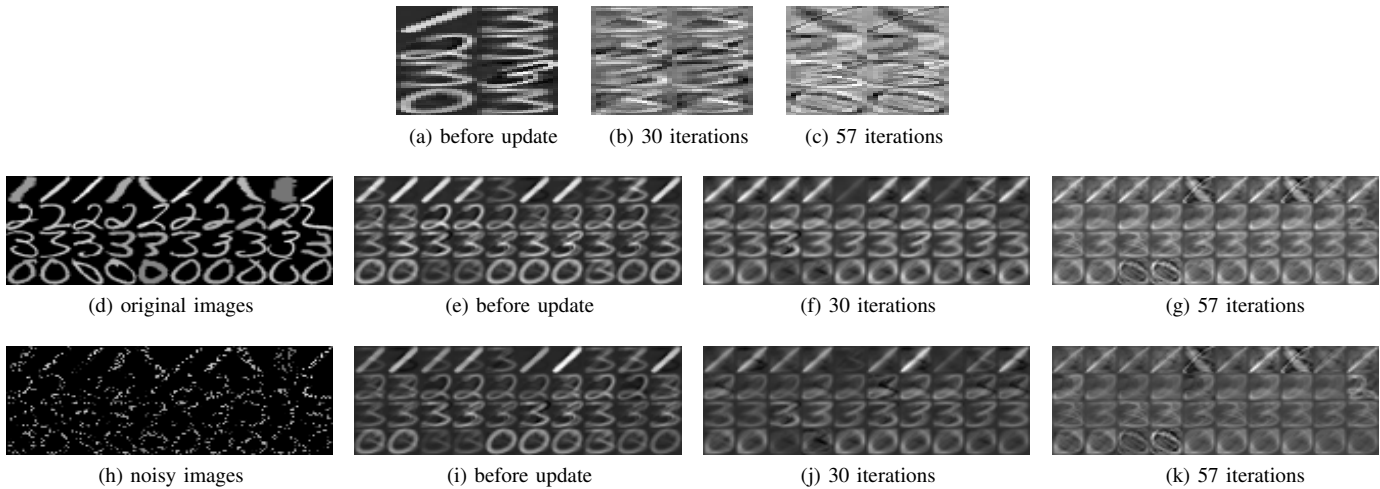

Fig. 5: Reconstruction using class dedicated atoms with the proposed dictionary update (sparsity 2 is used.). (a), (b) and (c) show the updated dictionary atoms, where from the top to the bottom the two atoms in each row are the dedicated atoms for class '1','2','3' and '0'. (e), (f) and (g) show the reconstruction to (d). (i), (j) and (k) show the reconstruction to (h). (h) are images in (d) with 60% missing pixels. Note that ITDU extracts the common internal structure of each class and eliminates the variation within the class, which leads to more accurate classification.

## V. CONCLUSION

We have presented an information theoretic approach to dictionary learning that seeks a dictionary that is compact, reconstructive and discriminative for the task of image classification. The algorithm consists of dictionary selection and update stages. In the selection stage, an objective function is maximized using a greedy procedure to select a set of compact, reconstructive and discriminative atoms from an initial dictionary. In the update stage, a gradient ascent algorithm based on the quadratic mutual information is adopted to enhance the selected dictionary for improved reconstruction and discrimination. Both the proposed dictionary selection and update methods can be easily applied for other dictionary learning schemes.

## ACKNOWLEDGMENT

The work was partially supported by a MURI from the Office of Naval Research under the Grant N00014-10-1-0934.

## REFERENCES

[1] R. Rubinstein, A. Bruckstein, and M. Elad, "Dictionaries for sparse representation modeling," *Proceedings of the IEEE*, vol. 98, no. 6, pp. 1045 –1057, Jun. 2010.

[2] J. Wright, Y. Ma, J. Mairal, G. Sapiro, T. Huang, and S. Yan, "Sparse representation for computer vision and pattern recognition," *Proceedings of the IEEE*, vol. 98, no. 6, pp. 1031 –1044, June 2010.

[3] M. Elad, M. Figueiredo, and Y. Ma, "On the role of sparse and redundant representations in image processing," *Proceedings of the IEEE*, vol. 98, no. 6, pp. 972 –982, June 2010.

[4] V. M. Patel and R. Chellappa, "Sparse representations, compressive sensing and dictionaries for pattern recognition," in *Asian Conference on Pattern Recognition (ACPR), Beijing, China*, 2011.

[5] S. Chen, D. Donoho, and M. Saunders, "Atomic decomposition by basis pursuit," *SIAM J. Sci. Comp.*, vol. 20, no. 1, pp. 33–61, 1998.

[6] Y. C. Pati, R. Rezaiifar, and P. S. Krishnaprasad, "Orthogonal matching pursuit: recursive function approximation with applications to wavelet decomposition," *Proc. 27th Asilomar Conference on Signals, Systems and Computers*, pp. 40–44, 1993.

[7] J. A. Tropp, "Greed is good: Algorithmic results for sparse approximation," *IEEE Trans. Info. Theory*, vol. 50, no. 10, pp. 2231–2242, Oct. 2004.

[8] B. A. Olshausen and D. J. Field, "Emergence of simple-cell receptive field properties by learning a sparse code for natural images," *Nature*, vol. 381, no. 6583, pp. 607–609, 1996.

[9] V. M. Patel, T. Wu, S. Biswas, P. J. Phillips, and R. Chellappa, "Dictionary-based face recognition under variable lighting and pose," *IEEE Transactions on Information Forensics and Security*, vol. 7, no. 3, pp. 954–965, June 2012.

[10] M. Aharon, M. Elad, and A. Bruckstein, "k-SVD: An algorithm for designing overcomplete dictionaries for sparse representation," *IEEE Trans. on Signal Processing*, vol. 54, no. 11, pp. 4311–4322, Nov. 2006.

[11] K. Etemand and R. Chellappa, "Separability-based multiscale basis selection and feature extraction for signal and image classification," *IEEE Trans. on Image Processing*, vol. 7, no. 10, pp. 1453–1465, Oct. 1998.

[12] M. Yang, X. F. L. Zhang, and D. Zhang, "Fisher discrimination dictionary learning for sparse representation," in *Proc. Intl. Conf. on Computer Vision, Barcelona, Spain*, 2011.

[13] F. Rodriguez and G. Sapiro, "Sparse representations for image classification: Learning discriminative and reconstructive non-parametric dictionaries," *Tech. Report, University of Minnesota*, Dec. 2007.

[14] E. Kokiopoulou and P. Frossard, "Semantic coding by supervised dimensionality reduction," *IEEE Trans. Multimedia*, vol. 10, no. 5, pp. 806–818, Aug. 2008.

[15] K. Huang and S. Aviyente, "Sparse representation for signal classification," in *Neural Information Processing Systems, Vancouver, Canada*, Dec. 2007.

[16] J. Mairal, F. Bach, and J. Ponce, "Task-driven dictionary learning," *IEEE TPAMI*, vol. 34, no. 4, pp. 791 –804, April 2012.

[17] J. Mairal, F. Bach, J. Pnce, G. Sapiro, and A. Zisserman, "Discriminative learned dictionaries for local image analysis," in *IEEE Computer Society Conf. on Computer Vision and Patt. Recn., Anchorage*, 2008.

[18] J. Mairal, F. Bach, J. Ponce, G. Sapiro, and A. Zisserman, "Supervised dictionary learning," in *Neural Information Processing Systems, Vancouver, Canada*, Dec. 2008.

[19] Q. Zhang and B. Li, "Discriminative k-SVD for dictionary learning in

face recognition," in *Proc. IEEE Computer Society Conf. on Computer Vision and Patt. Recn., San Francisco, CA*, June 2010.

[20] Z. Jiang, Z. Lin, and L. S. Davis, "Learning a discriminative dictionary for sparse coding via label consistent k-SVD," in *IEEE Computer Society Conf. on Computer Vision and Patt. Recn., Colorado Springs*, June 2011.

[21] Q. Qiu, Z. Jiang, and R. Chellappa, "Sparse dictionary-based representation and recognition of action attributes," in *Proc. Intl. Conf. Computer Vision, Barcelona, Spain*, Nov. 2011.

[22] J. A. Tropp, A. C. Gilbert, and M. J. Strauss, "Algorithms for simultaneous sparse approximation. part i: Greedy pursuit," *Signal Processing*, vol. 86, pp. 572–588, 2006.

[23] A. Krause, A. Singh, and C. Guestrin, "Near-optimal sensor placements in gaussian processes: Theory, efficient algorithms and empirical studies," *JMLR*, no. 9, pp. 235–284, 2008.

[24] M. E. Hellman and J. Raviv, "Probability of error, equivocation, and the Chernoff bound," *IEEE Trans. on Info. Theory*, vol. 16, pp. 368–372, 1979.

[25] C. M. Bishop, *Pattern Recognition and Machine Learning*. Springer, 2006.

[26] K. Torkkola, "Feature extraction by non parametric mutual information maximization," *JMLR*, vol. 3, pp. 1415–1438, Mar. 2003.

[27] J. Kapur, "Measures of information and their applications," 1994, wiley.

[28] A. S. Georghiades, P. N. Belhumeur, and D. J. Kriegman, "From few to many: Ilumination cone models for face recognition under variable lighting and pose," *IEEE Trans. Pattern Analysis and Machine Intelligence*, vol. 23, no. 6, pp. 643–660, June 2001.

[29] "USPS handwritten digit database." in *http://www-i6.informatik.rwth-aachen.de/ keysers/usps.html*.

[30] S. Lazebnik, C. Schmid, and J. Ponce, "Beyond bags of features: Spatial pyramid matching for recognizing natural scene categories," in *IEEE Computer Society Conf. on Computer Vision and Patt. Recn., New York, NY*, vol. 2, 2006, pp. 2169 – 2178.

[31] J. Yang, K. Yu, Y. Gong, and T. Huang, "Linear spatial pyramid matching using sparse coding for image classification," in *Proc. IEEE Computer Society Conf. on Computer Vision and Patt. Rec., Miami, FL*, June 2009.

[32] J. C. Gemert, J.-M. Geusebroek, C. J. Veenman, and A. W. Smeulders, "Kernel codebooks for scene categorization," in *Proc. European Conf. on Computer Vision, Marseiiles, France*, Oct. 2008.

[33] J. Wright, A. Yang, A. Ganesh, S. Sastry, and Y. Ma, "Robust face recognition via sparse representation," *IEEE TPAMI*, vol. 31, no. 2, pp. 210–227, 2009.

[34] J. Wang, J. Yang, K. Yu, F. Lv, T. Huang, and Y. Gong, "Locality-constrained linear coding for image classification," in *Proc. IEEE Computer Society Conf. on Computer Vision and Patt. Recn., San Francisco*, June 2010.